\documentclass[10pt,twocolumn,letterpaper]{article}

\usepackage[pagenumbers]{cvpr} 

\usepackage[accsupp]{axessibility} 

\pdfoutput=1
\usepackage[dvipsnames]{xcolor}

\definecolor{cvprblue}{rgb}{0.21,0.49,0.74}
\usepackage[pagebackref,breaklinks,colorlinks,citecolor=cvprblue]{hyperref}

\def\paperID{17550} 
\def\confName{CVPR}
\def\confYear{2024}

\title{U-VAP: \underline{U}ser-specified \underline{V}isual \underline{A}ppearance \underline{P}ersonalization via \\ Decoupled Self Augmentation}

\author{You Wu, Kean Liu, Xiaoyue Mi, Fan Tang$^{1*}$, Juan Cao, Jintao Li\\
Institute of Computing Technology, CAS\\
China\\
{\tt\small wuyou22s@ict.ac.cn, liukean22@mails.ucas.ac.cn, (tangfan,mixiaoyue19s,caojuan,jtli)@ict.ac.cn}
}

\begin{document}

\twocolumn[{
    \renewcommand\twocolumn[1][]{#1}
    \maketitle
    \begin{center}
    \centering
    \includegraphics[width=\linewidth]{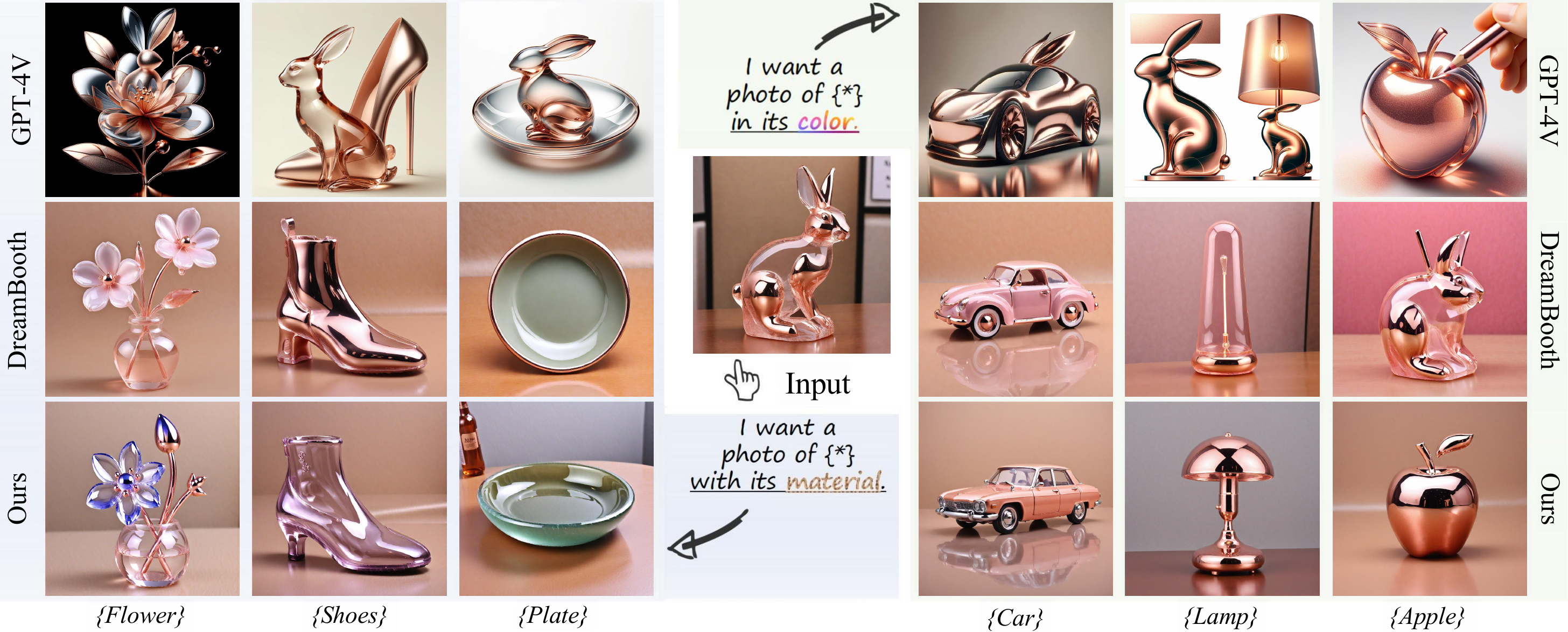}
    \captionof{figure}{Given a reference image,  
    U-VAP can personalize the user-specified visual appearances and combine them with some novel concepts.
    U-VAP generates images with the material (left) or color (right) extracted from the input image of a rabbit statue (middle) and achieves better accuracy and controllability in several new concepts.}
    \label{fig:teaser}
    \end{center}
}]

\renewcommand{\thefootnote}{}
\footnotetext{\textsuperscript{*}Corresponding author: Fan Tang.}

\begin{abstract}
Concept personalization methods enable large text-to-image models to learn specific subjects (e.g., objects/poses/3D models) and synthesize renditions in new contexts.
Given that the image references are highly biased towards visual attributes, state-of-the-art personalization models tend to overfit the whole subject and cannot disentangle visual characteristics in pixel space.
In this study, we proposed a more challenging setting, namely fine-grained visual appearance personalization.
Different from existing methods, we allow users to provide a sentence describing the desired attributes.
A novel decoupled self-augmentation strategy is proposed to generate target-related and non-target samples to learn user-specified visual attributes.
These augmented data allow for refining the model's understanding of the target attribute while mitigating the impact of unrelated attributes.
At the inference stage, adjustments are conducted on semantic space through the learned target and non-target embeddings to further enhance the disentanglement of target attributes.
Extensive experiments on various kinds of visual attributes with SOTA personalization methods show the ability of the proposed method to mimic target visual appearance in novel contexts, thus improving the controllability and flexibility of personalization.
Project page: \url{https://github.com/ICTMCG/U-VAP}.

\end{abstract}    
\section{Introduction}
\label{sec:intro}
Recent advancements in text-to-image models~\cite{rombach2022high,podell2023sdxl,saharia2022photorealistic,ramesh2022hierarchical} allow the generation of both fantastical and realistic high-resolution visual media from mere textual prompts.
To meet users' customized generation needs, personalization methods such as DreamBooth~\cite{ruiz2023dreambooth} have emerged, facilitating the learning of specific concepts from reference images and their integration with unique textual conditions.
Efforts are increasingly focused on enhancing the quality and efficiency of personalized specific concept~\cite{alaluf2023neural,gal2023encoder,han2023svdiff}, and combination of multiple concepts~\cite{kumari2023multi,avrahami2023break,zhang2023prospect,vinker2023concept}, significantly advancing artistic and creative applications. 

Despite these strides, fine-grained visual attributes (style/layout/texture) are still challenging to extract from few-shot image inputs, which are essential for new concept generation~\cite{zhang2023prospect}.
The primary difficulty lies in the inherent entanglement of these attributes within the same pixel space and the absence of explicit external supervision.
For instance, Fig.~\ref{fig:teaser} demonstrates the limitations of text-conditioned guidance in DreamBooth, where the lack of explicit supervision hinders the separation of specific visual attributes.
While recent works~\cite{sohn2023styledrop,wang2023stylediffusion,zhang2023inversion,wang2023styleadapter} have made progress in content and style disentanglement, achieving precise attribute separation remains an elusive goal.
Others rely on heuristic knowledge~\cite{voynov2023pplus,zhang2023prospect} or unsupervised learning~\cite{vinker2023concept}, lack of controllability.

In this paper, we propose a new setting for text-to-image model personalization, namely user-specified visual appearance personalization (U-VAP).
To learn specific visual attributes from limited reference images in a textually controllable way, U-VAP allows users to select desired visual attributes by giving text instructions.
An initial personalization model is first trained following DreamBooth~\cite{ruiz2023dreambooth}.
However, due to the input images often sharing the same visual attributes in the given samples, the initial personalization tends to couple all the visual appearances together and ignore the users' input query.

To this end, we propose a decoupled self-augmentation strategy.
Supported by the capabilities of advanced large language models~\cite{openaigpt35turbo,openai2023gpt4}, we generate two sets of instructions according to input prompts: one set containing the target attribute with enumerating other attributes, the other vice versa.
Then, the two sets of prompts are used to generate augmented samples by the initial personalization, which is used to further tune the models to couple target and non-target appearances.
In this way, U-VAP promotes the personalization towards user-specified attributes and away from irrelevant attributes.
Compared with SOTA personalization models, U-VAP can disentangle user-specified attributes accurately and flexibly combine them with other novel concepts.
The main contributions are as follows:
\begin{itemize}
\item[$\bullet$] We propose U-VAP, a simple yet effective approach for achieving user-specific visual appearance personalization from limited reference images.
  \item[$\bullet$] We construct a decoupled self-augmentation strategy by generating target and non-target augmentations to help generate the user-specified attributes accurately. 
  \item[$\bullet$] Experimental results demonstrate the effectiveness of U-VAP in various attribute-aware image generation tasks, which can be combined with other customized methods in a plug-and-play manner.
\end{itemize}

\section{Related Work}

\begin{figure*}[h]
    \centering
    \includegraphics[width=0.99\linewidth]{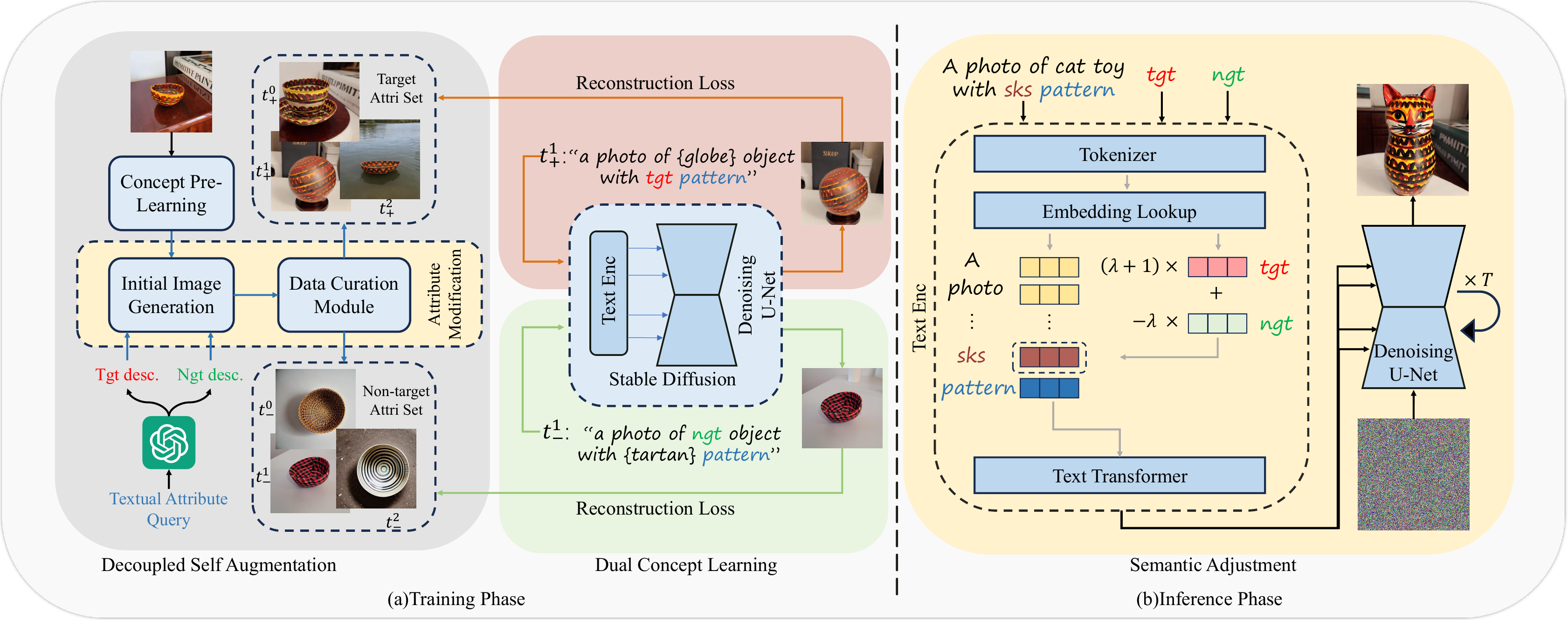}
    \caption{Pipeline of U-VAP. (a) Training: With input reference images, the initial concept-aware model is pre-trained on the entire concept.
    Meanwhile, given textual attribute query, U-VAP leverages GPT-3.5-turbo~\cite{openaigpt35turbo} to generate target and non-target descriptions for attribute modification.
    With these descriptions, U-VAP uses the initial concept-aware model to produce numerous candidate images and the data curation module filters them into target and non-target attribute sets.
    Subsequently, the identifiers $tgt$ and $ntg$ are optimized on each augmented set, which corresponds to the target and non-target attributes respectively.
    (b) Inference: we use semantic adjustment to correct the target embedding, further avoiding the entanglement of unwanted attributes in the generated results.}
    \label{fig:pipeline}
\end{figure*}  
\subsection{Personalization of diffution models}

The objective of personalization is to learn specific concepts from reference images and accurately reconstruct or freely edit them.
Based on the Text-to-Image diffusion model, Textual Inversion (TI)~\cite{gal2022TI} and Dreambooth~\cite{ruiz2023dreambooth} learn a single visual concept from several images by tuning the text encoder and entire model's parameters to bind the specific concept with a pseudo-words, respectively.

Some works~\cite{han2023svdiff,kumari2023multi,avrahami2023break,sohn2023styledrop,dong2022dreamartist,voynov2023pplus,zhang2023prospect,vinker2023concept} have explored methods to combine multiple concepts or attributes in personalization.
Custom Diffusion~\cite{kumari2023multi} allows for the joint training of multiple concepts or the combination of multiple fine-tuned models.
With the supervision of explicit masks on a single image and cross-attention maps, Break-a-Scene ~\cite{avrahami2023break} can learn multiple concepts separately.
However, they are unable to decouple specific attributes from a concept.
Different from previous concept-aware methods, Extended Textual Inversion (XTI)~\cite{voynov2023pplus} and ProSpect ~\cite{zhang2023prospect} discovered attribute-aware properties of diffusion model from neural structure and timestep spaces, which allows for flexible learning and combination of attributes from different concepts. 
They cannot explicitly decouple specified attributes but rely more on empirical adjustment.
Similar to attribute-aware personalization~\cite{voynov2023pplus,zhang2023prospect}, U-VAP aims to decouple user-specified attributes from reference concepts but enhances controllability by allowing textual guidance.

\subsection{Disentanglement in Generative Models}

Previous studies have demonstrated that pre-trained GAN s~\cite{brock2018large,karras2021alias,karras2019style,karras2020analyzing}  can achieve disentanglement by traversing specific directions in their latent space ~\cite{harkonen2020ganspace,shen2020interpreting,shen2020interfacegan,shen2021closed,li2022contrastive,deng2020disentangled} to resulting in changes on specific attributes~\cite{abdal2019image2stylegan,abdal2020image2stylegan++,patashnik2021styleclip,gal2022stylegan, huang2024controllable}. 

Recently there were some works exploring disentanglement in diffusion models~\cite{kwon2022diffusion,preechakul2022diffusion,wu2023uncovering,sohn2023styledrop,dong2022dreamartist,wang2023styleadapter,wang2023stylediffusion}.
~\citet{kwon2022diffusion} and ~\citet{preechakul2022diffusion} explore the disentanglement with the aspect of latent representation modification for image generation and attribute transfer. 
~\citet{wu2023uncovering} optimize the mixing weights of the source and target text embeddings for disentangled image editing. 
Recently, advanced image editing methods~\cite{hertz2022prompt,brooks2023instructpix2pix,mokady2023null,miyake2023negative} achieve more accurate attribute modification textual controllably.
However, these editing methods have limitations in the diversity and flexibility of specified attribute modification.
Several works ~\cite{wang2023styleadapter,zhang2022domain,wang2023stylediffusion,zhang2023inversion,sohn2023styledrop,dong2022dreamartist} have been proposed for extracting the specific style from reference images, which can be regarded as a coarse disentanglement only between content and style.
~\citet{voynov2023pplus} and ~\citet{zhang2023prospect} heuristically choose the embeddings obtained from neural or time spaces based on the prior knowledge of frequency characteristics of different visual attributes, which makes them difficult to distinguish attributes with similar characteristics.
~\citet{vinker2023concept} treats a single concept as a composition of multiple concepts and offers a concise but uncontrollable method for decomposition.
By constructing attribute-aware samples via the self-augmentation strategy, U-VAP has the capability to controllably and flexibly personalize the specified attributes from an entire concept. 

\section{Method}

Based on the Stable Diffusion (SD) model~\cite{rombach2022high}, we construct our personalization framework U-VAP on the foundation of a basic technique, DreamBooth~\cite{ruiz2023dreambooth}.
User input a small set of reference images $I_{ref}$ containing multiple attributes $a= \left\{a_{tgt}, a_{ngt}\right\}$, where $a_{tgt}$ is a target attribute provided by user input prompt (e.g. ``object’s structure''), and $a_{ngt}$ is the other non-target attributes in reference images $I_{ref}$.
Our objective is to accurately extract a particular target attribute $a_{tgt}$ and apply it to the creation images $I_{output}$ of new concepts using new text prompt $y$. 
It can be formalized as:
\begin{equation}
  \begin{aligned} 
  I_{output} = G(a_{tgt},y,I_{ref}),
  \end{aligned}
  \label{eq:task_formalization}
\end{equation}
where $G$ refers to a model used for generating images based on specific attributes.
For this purpose, we extend the general personalization process and propose a framework.
As shown in Fig.~\ref{fig:pipeline}, users' input reference images of a concept and a textual attribute query are input data.
U-VAP first learns the initial concept-aware model using DreamBooth (Sec.~\ref{subsec:DreamBooth}).
Simultaneously, the decoupled self-augmentation module (Sec.~\ref{subsec:CAA}) in U-VAP utilizes a large language model (LLM) to generate target and non-target attribute descriptions for attribute modification.
Using these descriptions, the concept-aware model generates numerous candidate images. These are then filtered into target and non-target attribute sets.
Finally, identifiers $tgt$ and $ngt$ are optimized for the augmented sets, corresponding to the $a_{tgt}$ and $a_{ngt}$, respectively.
In the inference time, semantic adjustment is employed to refine the target embedding (Sec.~\ref{subsec:adjustment}).
This step is important in preventing the entanglement of non-target attributes in the generated images.

\subsection{Preliminaries for Initial Model}
\label{subsec:DreamBooth}
In the concept pre-learning step, we leverage DreamBooth~\cite{ruiz2023dreambooth} to pre-learn the entire concept and obtain a concept-aware personalization model $G_0$ first.
DreamBooth is a subject-driven personalization method, which directly fine-tunes the parameters of the diffusion model with reference images $I_{ref}$. 
It identifies the abstract categories of the subject and then constructs the text prompts using a unique identifier followed by the class name(e.g., ``A sks dog'').
With text prompts and reference images, DreamBooth conducts fine-tuning with the same reconstruction objective as SD.
To prevent language drift~\cite{lee2019countering,lu2020countering}, they propose a class-specific prior preservation loss.
The parameters of the model are optimized simultaneously on both the reference images $I_{ref}$ and class images generated by its original version, conditioned by text prompt and class name (e.g., A ``dog'') respectively:
\begin{equation}
  \begin{aligned}  \min _\theta \mathbb{E}_{z, y, \varepsilon, \varepsilon^\prime, t}\left[\left\|\varepsilon - \varepsilon_\theta\left(z_t, t, c(y)\right)\right\|_2^2 \right. \\ \left. + \alpha \left\|\varepsilon^\prime - \varepsilon_\theta\left(z_t^\prime, t, c(y^\prime)\right)\right\|_2^2 \right]. \end{aligned}
  \label{eq:2}
\end{equation}
Here the latent code $z\sim\mathcal{E}(x)$ represents encoded input image $I_{ref}$, and $\mathcal{E}$ is the image encoder of SD. 
$z_t$ and $z_t^\prime$ are latent codes of user-provided image and class image by adding noise $\varepsilon \sim \mathcal{N}(0,1)$ to $z$ at timestep $t$.
$y$ is the constructed text prompt and $y^\prime$ is the class name for class-specific prior preservation.
$c(y)$ is conditioning vector, and $c$ is the text encoder of SD.
Given noised latent $z_t$, the timestep $t$, and conditioning $c(y)$, the denoising U-Net $\varepsilon_\theta$ predicts and removes the added noise.
$\alpha$ represents the weight of class-specific prior preservation loss.


\begin{figure}[!t]
    \centering
    \includegraphics[width=0.99\linewidth]{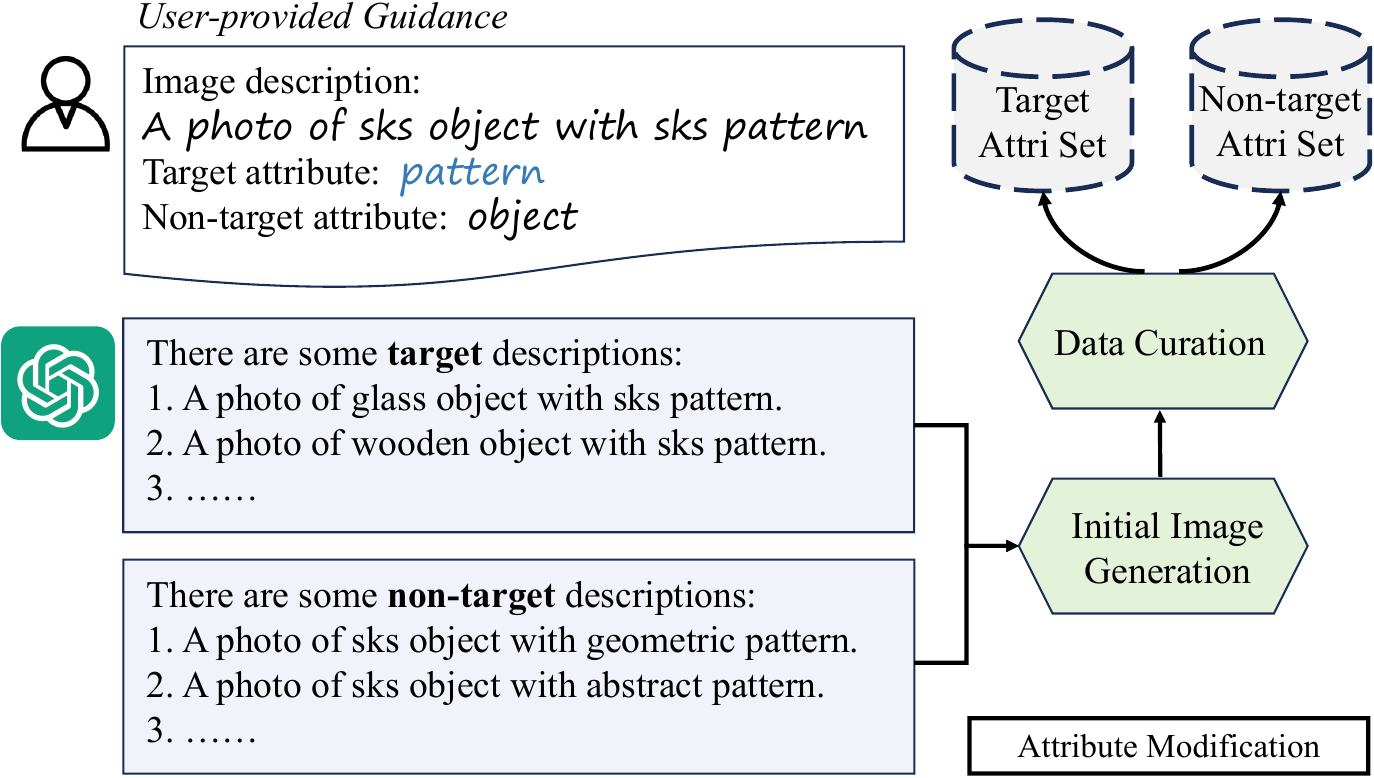}
    \caption{The workflow for decoupled self-augmentation. 
    The ``sks'' represents the initial identifier learned in pre-learning step.
    With transformed query from guidance, an LLM~\cite{openaigpt35turbo} modifies the attribute descriptors and generates target and non-target descriptions respectively.
    Then the initial concept-aware model produces numerous candidate images. 
    After data curation, U-VAP constructs attribute-aware samples for dual-concept learning.
    }
    \label{fig:ASC}
\end{figure}

\subsection{Learning via Decoupled Self Augmentation} 
\label{subsec:lll}
The lack of attribute-aware data is a primary obstacle in achieving precise personalization of specific attributes. 
Based on that, we explore a decoupled self-augmentation strategy by constructing two sets of attribute-aware samples using a large language model (LLM).
Specifically, we create a target attribute text prompt set $T_{+}$ by preserving the target attributes $a_{tgt}$ (e.g. pattern) and modifying the undesired non-target attributes $a_{ngt}$ (e.g. object's structure).
We also construct a non-target attribute text prompt set $T_{-}$ by modifying only $a_{tgt}$.
Then we generate two sets of images $I_{+}$ and $I_{-}$ by initial concept-aware model $G_{0}$ according to $T_{+}$ and $T_{-}$,respectively. 
Finally, we select images from $I_{+}$ and $I_{-}$ that meet the criteria for constructing attribute-aware samples in the data curation module.

\paragraph{LLM-Powered Prompt Generation.} 
\label{subsec:CAA}
To generate a diverse set of samples, U-VAP leverages an LLM (Specifically, GPT-3.5-turbo~\cite{openaigpt35turbo}) to automatically generate a large number of text prompts.
Specifically, users first provide U-VAP with textual guidance.
Then U-VAP transforms it into a query and sends it to the LLM. 
As shown in Fig.~\ref{fig:ASC}, the guidance should offer an initial description of the user's input concepts to determine the template of text prompts.
By specifying $a_{tgt}$ and $a_{ngt}$, LLM can construct target descriptions $T_{+}$ and non-target descriptions $T_{-}$ as text prompts respectively.

\begin{figure}[!t]
    \centering
    \includegraphics[width=\linewidth]{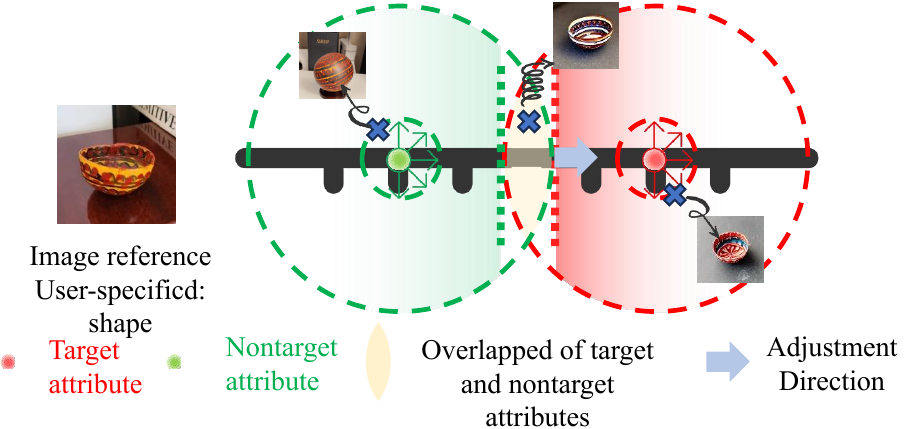}
    \caption{Illustrations for semantic adjustment. By shifting semantic embedding on the adjustment direction, U-VAP can further promote the elimination of non-target attributes in the results.}
    \label{fig:adjust}
\end{figure}

\paragraph{Initial Image Generation.}
$G_0$ associates the entire concept with a unique identifier $sks$ after pre-learning of it. 
Using this identifier followed by an attribute word in the constructed text prompt (e.g., ``A photo of $sks$ structure backpack''), $G_0$ attempts to integrate this attribute onto certain new concepts (e.g., ``backpack'').
With these descriptions, U-VAP generates images $I_{+}=G_0(T_{+})$ and $I_{-}= G_0(T_{-})$.

\begin{figure*}[h]
    \centering
    \includegraphics[width=0.95\linewidth]{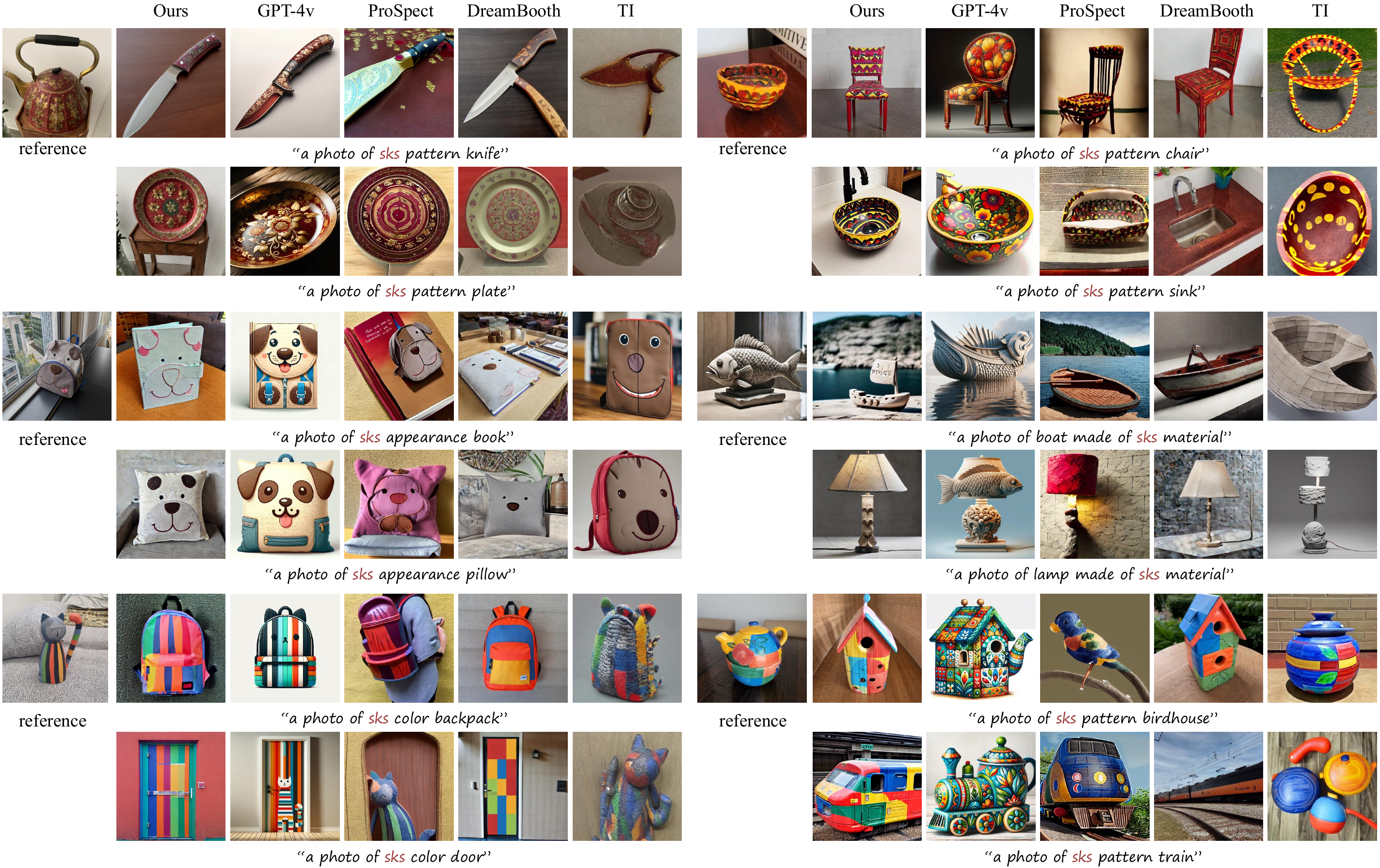}
    \caption{Qualitative comparisons. Compared with SOTA personalization methods GPT-4V, ProSpect, DreamBooth and TI, we can achieve controlled and precise generation of specific visual attributes while maintaining high visual quality.}
    \label{fig:main_result2}
\end{figure*}

\paragraph{Data Curation.}
Due to the limited effectiveness of attribute modification in concept-level personalized approaches, the generated samples often fail to meet our requirements.
It is necessary to filter $I_{+}$ and $I_{-}$ to obtain the final attribute-aware samples $D=\left\{ D_{+}, D_{-} \right\}$, represented as $D_{+}=\left\{ T_{+}, I_{+}\right\}$ and $D_{-}=\left\{ T_{-}, I_{-}\right\}$.
Affected by the new attribute (style or object) in the target and non-target descriptions, many images from the initial samples lose crucial information in the references.
Consequently, we employ CLIP image similarity with the original image as a metric to automatically sample images with higher similarity.
This criterion could at least guarantee that the filtered samples are relatively faithful to the original references, and ensures the learning process does not make the model perform worse.
For introducing human intention and constructing better attribute-aware samples $D$, we strongly suggest utilizing Human Feedback for further selection after automatic filtering.
Finally, we respectively sample $m$ images for $D_{+}$ and $D_{-}$.

\paragraph{Dual Concept Learning.} 
U-VAP simultaneously conducts personalization on $D_{+}$ and $D_{+}$.
As illustrated in Fig.~\ref{fig:pipeline}, the training process has two branches based on different attribute sets and learns two identifiers, $tgt$, and $ngt$.
Taking the target attribute set $D_{+}=\{ T_{+}, I_{+}\}$ as an example, where $T_{+}=\left\{ t_{+}^0, t_{+}^1, \cdots, t_{+}^m \right\}$ and $I_{+}=\left\{ x_{+}^0, x_{+}^1, \cdots, x_{+}^m \right\}$, $x_{+}^i$ corresponds to a target description $t_{+}^i$ and we replace the  identifier $sks$ with target identifier $tgt$ (e.g., ``a photo of a globe object with a tgt pattern'').
When conditioned by $t_{+}^i$, the model is optimized to reconstruct $x_{+}^i$. 
Ultimately the identifier $tgt$ is bound to the concept described in $D_{+}$.
Due to the augmentation on the target attribute, $tgt$ is promoted to be optimized towards the semantic direction of the target attribute.

The objective of U-VAP is:
\begin{equation}
  \begin{aligned}  \min _\theta \mathbb{E}_{z, y, \varepsilon, \varepsilon^\prime, t}\left[\left\|\varepsilon - \varepsilon_\theta\left(z_t^{+}, t, c(t_{+})\right)\right\|_2^2 \right. \\ \left. + \left\|\varepsilon^\prime - \varepsilon_\theta\left(z_t^-, t, c(t_{-})\right)\right\|_2^2 \right] , \end{aligned}
  \label{eq:2}
\end{equation}
where $\theta$ is the fine-tuning parameters of the model, $z_t^{+}$ and $z_t^{-}$ are latent codes of $I_{+}$ and $I_{-}$, respectively.
Different from vanilla DreamBooth, we do not consider the class-specific prior preservation loss.
After dual concept learning, $tgt$ and $ngt$ independently acquired visual information related to the target attributes and non-target attributes.

\begin{figure*}[t]
    \centering
    \includegraphics[width=0.90\linewidth]{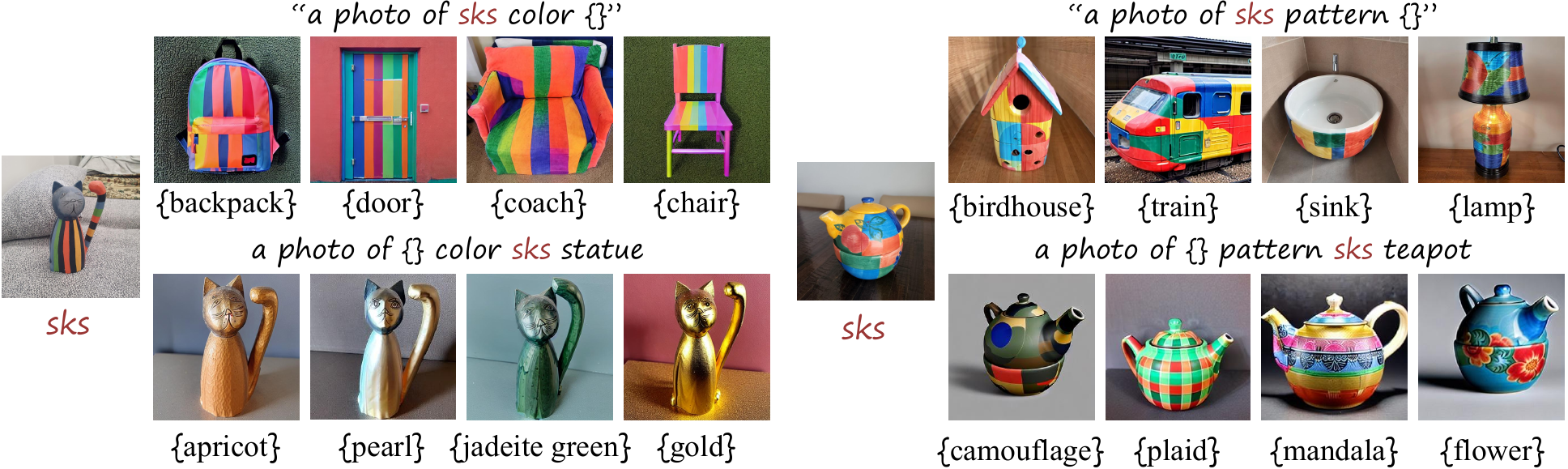}
    \caption{Personalization of different attributes from the same reference images. Our method allows precise customization of multiple attributes~(e.g., color and statue in the figure) of the reference image, as well as generalization of other new concept combinations.}
    \label{fig:main_result1}
\end{figure*}
\subsection{Semantic Adjustment} 
\label{subsec:adjustment}

Fig.~\ref{fig:adjust} illustrates overlapping areas in semantic space for $tgt$ and $ngt$ embeddings. 
As a result, non-target attributes may be present in generated results.
To tackle this issue, we propose a semantic adjustment during the inference phase.
Starting from $tgt$, we shift the embedding representation away from $ngt$ in the semantic space as the correction.
In Fig.~\ref{fig:pipeline}, we introduce a placeholder $sks$ for $tgt$ to compose the text prompt $p$.
Prior to feeding the text embedding $V$ of $p$ into the text transformer, the vector corresponding to $sks$ is replaced according to:
\begin{equation}
  v^{sks}=v^{tgt} + \lambda \overrightarrow{(v^{tgt} - v^{ngt})} ,
  \label{eq:adjust}
\end{equation}
where $v^{sks}$, $v^{tgt}$ and $v^{ngt}$ are embedding vectors of $sks$, $tgt$ and $ngt$.
$\lambda$ represents the degree to which $v^{tgt}$ is shifted in the direction of $v^{tgt} - v^{ngt}$. 
Due to the varying severity of non-target attribute entanglement at different sampling noises as starting points or concepts combined, the optimal value of $\lambda$ is also difficult to fix.
Considering the cost of parameter adjustment, we choose to apply this operation at the inference phase.

\section{Experiments}

In this section, we demonstrate the effectiveness of our method in controlling personalization on various visual attributes.
We conduct qualitative and quantitative experiments in Sec.~\ref{sec:main_results}, and user studies in the supplementary materials.
Finally, we show sufficient ablation results in Sec.~\ref{sec:abla_results}.

\subsection{Experimental Setting}
\paragraph{Training Details.} 
Our experiments are based on Stable Diffusion v1.5~\cite{rombach2022high} with default hyperparameters.
For data curation, we automatically filter $10\%$ from approximately $200$ candidate images, and $2 \sim 6$ training images will be selected by human feedback (equivalent to the number of baselines). 
We set a learning rate of 5e-6 with 500 optimization steps at pre-learning stage.
In the training phase with attribute-aware sets, we set the same learning rate with 1000 optimization steps to learn the target and non-target attributes.
At the inference stage, we employed the diffusion steps in DDIM sampler of $T = 50$, and a guidance scale of $7.5$.
The synthesis process takes about 5 seconds, depending on the number of diffusion steps taken.
All experiments are based on NVIDIA A100 with a batch size of 1.

\paragraph{Baselines.} We choose Textual Inversion (TI)~\cite{gal2022TI}, DreamBooth (DB)~\cite{ruiz2023dreambooth}, GPT-4V~\cite{openai2023gpt4v}, and ProSpect~\cite{zhang2023prospect} as our baselines to demonstrate our effectiveness in specific appearance personalization.
(1)~TI learns a pseudo-word for a concept within a limited number of images using an optimization-based approach.
(2)~DB is a widely used method that learns a unique identifier and fine-tunes the diffusion model to learn the concept from a set of images.
(3)~GPT-4V is a state-of-the-art muli-modal language model, which can understand texts and images and use DALL·E 3~\cite{dalle3} as a tool to achieve personalization generation.
(4)~ProSpect learns and combines different identifiers in the expanded textual conditioning space to empirically achieve attribute personalization.
Our method can combine with TI and DB in a plug-and-play manner.

\begin{figure}[t]
    \centering
    \includegraphics[width=0.95\linewidth]{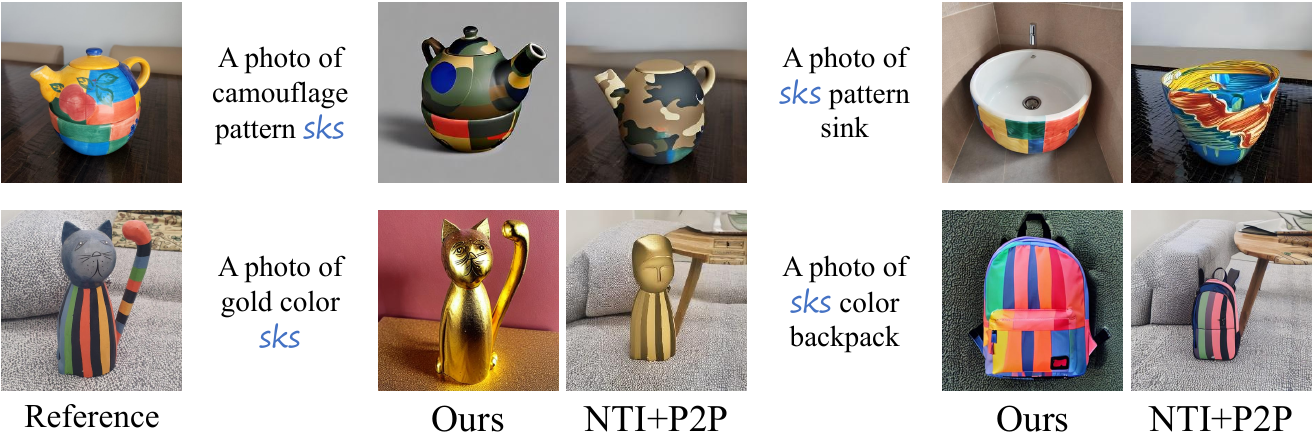}
    \caption{Comparison with NTI~\cite{mokady2023null}+P2P~\cite{hertz2022prompt}. The 2nd and the 5th columns are text conditions, and the 2 columns on their right are generated images by our method and NTI+P2P. ``sks'' is the identifier of the specific visual attribute.}
    \label{fig:P2P}
\end{figure}

\paragraph{Evaluation Metrics.}
We report quantitative results based on CLIP~\cite{radford2021learning} and Inception Score(IS)~\cite{salimans2016IS} to evaluate both prompt fidelity, image fidelity and generation quality.
Prompt fidelity is measured as the average CLIP cosine similarity between the generated image and prompt condition, which is named CLIP-T.
Image fidelity is the average CLIP cosine similarity between the generated and reference images, which is named CLIP-I.
Inception Score is calculated to evaluate the quality and diversity of generated images.
Furthermore, we execute a study focused on user preferences to further assess these aspects in the supplementary materials.
\subsection{Main Results}
\label{sec:main_results}
\paragraph{Quantitative Evaluation} 
We generated 1272 images and divided target attributes into three categories: color, pattern (non-target attribute: object), and structure (like object, shape, etc., non-target attribute: color or pattern).
Then we computed prompt fidelity and Inception Score~\cite{salimans2016IS} to evalute our method with TI~\cite{gal2022TI}, DB~\cite{ruiz2023dreambooth} and ProSpect~\cite{zhang2023prospect}.
As Tab.~\ref{tab:quantity} shown, when learning ``color'' or ``pattern'', our method and DB exhibit the highest CLIP-T and Inception Score (IS).
DreamBooth often excessively edits towards the new concept when generating with some attributes, such as color, material, and pattern (the train in the 6th row and the 3rd column from the right in Fig.~\ref{fig:main_result2}).
Therefore, CLIP-T cannot fully illustrate DB's editing accuracy. 
Excessive editing also results in increased generation quality and diversity, reflected in the highest IS score.
TI~\cite{gal2022TI} often results in entanglement between non-target attributes and new concepts' attributes in images.
ProSpect~\cite{zhang2023prospect} is capable of combining attributes in some cases, but it encounters difficulties in decoupling attributes that have similar frequency characteristics.
For CLIP-I, DB shows the lowest score followed by ours because it usually over-edits the non-target attributes. 
While Textual Inversion (TI) and ProSpect usually have higher CLIP-I, because they often face challenges in modifying objects.
When learning ``structure'', DB encounters the same challenges as TI and ProSpect.
Our method achieves lower CLIP-I than DB, but has the highest CLIP-T because it better combines the reference structure with some new appearances.


\paragraph{Qualitative Evaluation} 

As shown in Fig.~\ref{fig:main_result2}, we compare our method with four SOTA personalization methods TI~\cite{gal2022TI}, DreamBooth~\cite{ruiz2023dreambooth}, ProSpect~\cite{zhang2023prospect}, and GPT-4V~\cite{openai2023gpt4v}.
To conduct a fair and unbiased evaluation, we use cases from previous papers.
We augment each set of images with different texts to demonstrate the flexibility of our method.
We observe that GPT-4V struggles to maintain precise target attributes. For instance, in the third-row case, GPT-4V learned a cartoon-style appearance for the dog, which was inconsistent with the realistic style of the reference image, and the details of the dog (such as the eyes and mouth) do not align with the reference.
In contrast, our method successfully produces book and pillow images in a realistic style while preserving more details.
ProSpect and DB show inferior performance in integrating target attributes with added semantic content.
Although ProSpect maintains the target attributes, the visual quality of its generated chairs and sinks are poor.
DreamBooth, on the other hand, produces a chair and sink of high quality but lacks consistency in target attributes.
Our method, however, can generate these concepts with high-quality reference art patterns.
TI exhibits poor performance in both consistency of target attributes and in adding semantic content (e.g., chair and sink), and sometimes overfits to the reference, leading to a lack of editability, as in the first case where it ignores textual input and generates a pot-like shape.
Our approach effectively maintains target attributes while generating images with richer semantic content.

Moreover, we demonstrate the effectiveness of our approach in personalization generation of specific appearances, by showcasing customized results for different attributes using the same reference image.
As illustrated in Fig.~\ref{fig:main_result1}, our study presents customized generation results for different attributes using the same reference image, thereby demonstrating its efficacy in controlled attribute decoupling. In Fig.~\ref{fig:main_result1}, a colored cat statue image is used as the reference, where we have performed precise and controllable customization on both its color appearance (the first line) and statue structure (the second line).
By maintaining the color appearance, our method can generate diverse objects such as multicolored backpacks, doors, coaches, and chairs.
Furthermore, by preserving the ``statue'', we can produce toy cats in hues of apricot, pearl, jadeite green, and gold.
Those cases highlight the ability of our method to modify and control distinct attributes within an image selectively.

In Fig.~\ref{fig:P2P}, we compare our method with Null-text Inversion (NTI)~\cite{mokady2023null} + Prompt-to-Prompt (P2P)~\cite{hertz2022prompt}, which is a diffusion-based text-guidance image editing method.
NTI+P2P's editing is considerably constrained by the basic structure and layout of the original image, resulting in edited results that do not meet the textual requirements.
For example, the backpack with the color of the cat statue generated by NTI+P2P has the same structure as the reference.
In contrast to NTI+P2P, our approach exhibits better editability and generates images with superior diversity. 

\begin{table}[!t]
\vspace{-10pt}
\setlength{\abovecaptionskip}{0.1cm}
\centering
\caption{Quantitative evaluations. The highest results are in bold, and the second highest results are underlined.}
\resizebox{0.99\linewidth}{!}{
\begin{tabular}{lccc|ccc|ccc}
\hline Attributes & & Color & & & Pattern & & & Structure & \\
\hline Method & CLIP-I $\uparrow$  & CLIP-T  $\uparrow$ & IS     $\uparrow$ & CLIP-I $\uparrow$  & CLIP-T  $\uparrow$ & IS     $\uparrow$ & CLIP-I $\uparrow$  & CLIP-T  $\uparrow$ & IS     $\uparrow$ \\
\hline 
Ours & 0.497   & $\underline{0.292}$ & $\mathbf{4.138}$                & 0.470   & $\underline{0.270}$ & $\underline{3.434}$             & 0.709   & $\mathbf{0.284}$    & $\underline{1.740}$  \\
ProSpect & $\underline{0.581}$ & 0.239 & 2.691                             & $\underline{0.567}$ & 0.261 & 3.364                             & $\underline{0.773}$ & 0.259 & 1.258  \\
DreamBooth & 0.471&$\mathbf{0.303}$&$\underline{3.890}$&0.467&$\mathbf{0.299}$&$\mathbf{3.793}$&$\mathbf{0.837}$&$\underline{0.265}$&1.520   \\
TI & $\mathbf{0.608}$ & 0.187 & 2.818                           & $\mathbf{0.641}$ & 0.152 & 2.538                              & 0.618 & 0.191 & $\mathbf{2.434}$   \\
\hline
\end{tabular}
}
\label{tab:quantity}
\vspace{-10pt}
\end{table}

\begin{figure*}[t]
    \centering
    \includegraphics[width=0.90\linewidth]{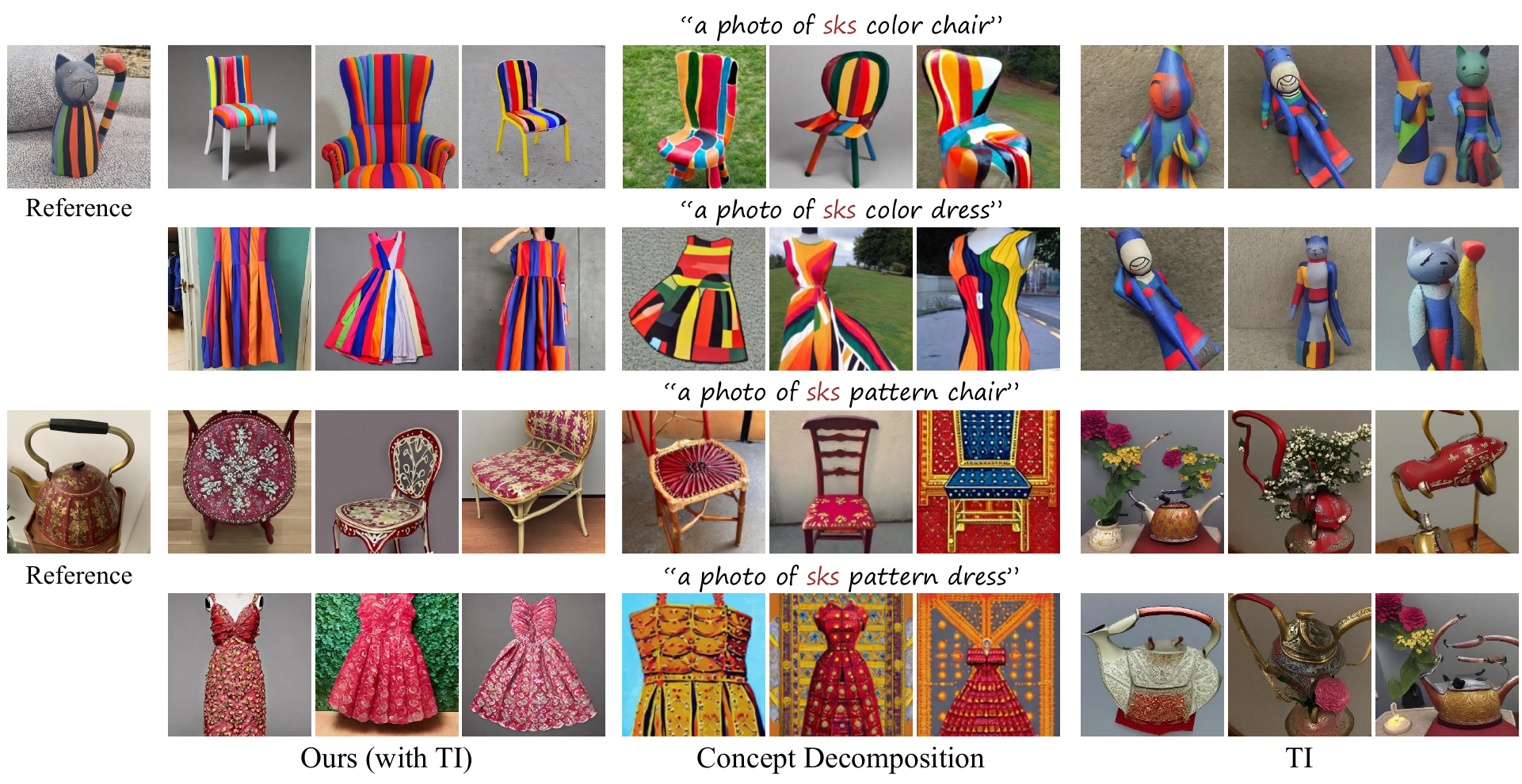}
    \caption{Comparison with TI~\cite{gal2022TI} and Concept Decomposition~\cite{vinker2023concept}. Our method performs better in generating specified attributes.}
    \label{fig:TIours}
\end{figure*}

Besides, we demonstrate that our proposed framework can also enhance the attribute personalization capabilities of Textual Inversion (TI)~\cite{gal2022TI}.
Unlike DreamBooth~\cite{ruiz2023dreambooth}, TI fine-tunes the textual embedding space of SD's text encoder to learn an identifier representing a specific concept.
We can deploy TI into U-VAP by optimizing target and non-target identifiers' embedding vectors.
In Fig.~\ref{fig:TIours} we compare our method with TI and Concept Decomposition~\cite{vinker2023concept}.
As shown in Fig.~\ref{fig:TIours}, conditioned by ``a photo of sks color chair'' and ``a photo of sks color dress'' with the reference image, TI is almost unable to generate the color personalized images.
Compared with Concept Decomposition, our method generates more accurate images of chairs and dresses, and preserves the specified color effectively.
It indicates that our method can significantly improve the TI's attribute personalization ability.
Similar to Concept Decomposition~\cite{vinker2023concept}, U-VAP is able to combine different visual attributes (as shown in the supplementary materials).

\begin{figure}[t]
    \centering
     \includegraphics[width=0.75\linewidth]{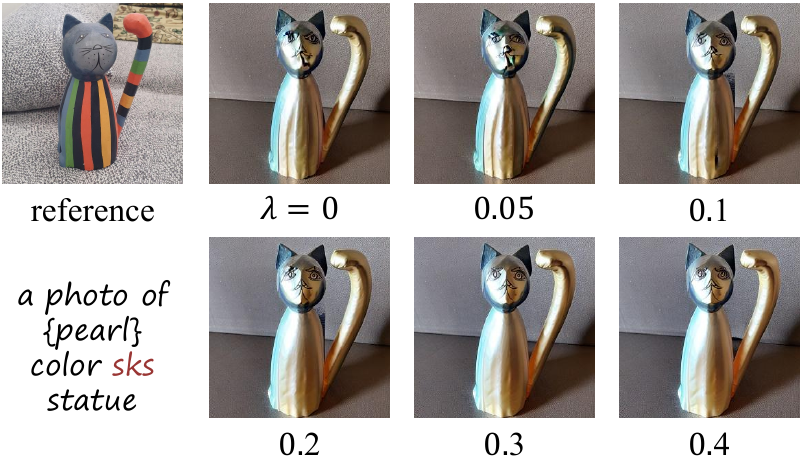}
    \caption{Ablation of $\lambda$ in semantic adjustment. 
    With $\lambda$ increasing, the quality and semantic accuracy are improved.}
    \label{fig:ablation}
\end{figure}

\subsection{Ablations}
\label{sec:abla_results}

At the inference phase, the $\lambda$ in Eq.~\ref{eq:adjust} of semantic adjustment affects the accuracy and quality of generation. 
Fig.~\ref{fig:ablation} demonstrates the influence of different values of $\lambda$ when generating a new concept with the same structure as the reference.
when $\lambda=0$, semantic adjustment is not applied and leads to the appearance of undesirable non-target attributes in results, such as the stripe-like pattern.
Additionally, structural details in the generated images are flawed, including distorted facial features and non-smooth edges, contributing to an overall decrease in quality. 
We believe that it is due to the insufficient disentanglement of target attributes learned from the target attribute set, resulting in many non-target attributes remaining.
As $\lambda$ increases, the embedding vector of the target identifier $tgt$ is corrected towards a direction away from non-target attributes. 
When $\lambda \geq 0.3$, the visual content of unrelated attributes in the generated images is significantly reduced. 
For all results without additional explanation in the paper, we set $\lambda=0.3$.

Furthermore, the number of constructed target and non-target attribute sets also affects the personalization. More discussions are shown in the supplementary materials.


\section{Conclusion}
In this paper, we analyzed the reasons for the difficulty in specific visual appearance personalization.
From the perspective of attribute-aware sample construction, we introduce a novel personalization framework named U-VAP.
U-VAP can learn the user-specified attributes controllably and generate high-quality results with textual guidance.
Extensive experiments indicate that existing works can be flexibly deployed into our framework and are improved effectively in attribute extracting.
We plan to further analyze better self-augmentation methods in attribute disentanglement for making more accurate personalization in the future.

\noindent \paragraph{Limitations.}
Our method's dependence on the capability of basic concept-aware personalization methods (such as DreamBooth~\cite{ruiz2023dreambooth}) results in poor performance in some attributes' decoupling during pre-learning.
For more discussions, please see supplementary materials.

%


\section*{Acknowledgements}
This work was partly supported by the National Natural Science Foundation of China under No. 62102162, Beijing Science and Technology Plan Project under No. Z231100005923033.

{
    \small
    \bibliographystyle{ieeenat_fullname}
    \bibliography{main}
}

\clearpage
\setcounter{page}{1}
\renewcommand\thesection{\Alph{section}}
\renewcommand{\thetable}{S\arabic{table}}
\renewcommand{\thefigure}{S\arabic{figure}}
\maketitlesupplementary

In Sec.~\ref{sec:userstudy}, we provide user studies for human preference evaluation.
We demonstrate the ability of attribute combination in Sec.~\ref{sec:combination} and analyze the impact of constructed attribute-aware samples on personalization in Sec.~\ref{sec:Data}.
The limitation of our method is discussed in Sec.~\ref{sec:limitation}.

\section{User Study}
\label{sec:userstudy}
We conduct user studies to evaluate the proposed U-VAP in specified visual appearance personalization on seven different concepts with three baseline methods: Textual Inversion~(TI)~\cite{gal2022TI}, DreamBooth~\cite{ruiz2023dreambooth}, and ProSpect~\cite{zhang2023prospect}. 
To obtain one set of image results, given a reference concept and the specified attribute, four images of a certain novel concept are generated by each method. 
Every volunteer is asked to select the best image based on two criteria to evaluate per image set:

\begin{itemize}[leftmargin=0.5cm, itemindent=0.2cm]
    \item Criterion-\uppercase\expandafter{\romannumeral1}: the accuracy of attribute personalization. 
    \item Criterion-\uppercase\expandafter{\romannumeral2}: the quality of novel concept generation on the premise of attribute accuracy.
\end{itemize}

We generated 24 sets of these images using different specified attributes and concepts, and a total of 55 volunteers participated in our study.
As shown in Tab.~\ref{tab:userstudy}, when evaluating the highest accuracy of attribute personalization, the images from our method were voted with a probability of $64.71\%$.
The results by DreamBooth were chosen with the second maximum probability, which was only $15.42\%$
Taking consideration of attribute accuracy, our images were voted with a probability of $53.84\%$ for choosing the best quality of novel concept generation.
However, the maximum probability of selecting images from another method is only $17.92\%$ (TI).
Compared with the other three baseline methods, images generated by our method were chosen with higher probability.
These results indicate U-VAP generates novel concepts with accurately specified attributes, exhibiting better performance in human preference.
\begin{table}[h!]
\centering
\caption{User study on two criteria. Criterion-\uppercase\expandafter{\romannumeral1}: The accuracy of attribute personalization. Criterion-\uppercase\expandafter{\romannumeral2}: The quality of novel concept generation. The highest voting probability is in bold.}
\resizebox{0.7\linewidth}{!}{
\begin{tabular}{lcc}
\hline Methods & Criterion-\uppercase\expandafter{\romannumeral1}     & Criterion-\uppercase\expandafter{\romannumeral2}   \\
\hline 
Ours       & $\mathbf{67.50\%}$   & $\mathbf{57.22\%}$ \\
ProSpect   & $8.47\%$                  & $10.69\%$ \\
DreamBooth & $15.42\%$                  & $14.58\%$ \\
TI         & $8.47\%$                  & $17.92\%$ \\
\hline
\end{tabular}
}
\label{tab:userstudy}
\end{table}

\section{Attribute Combination}
\label{sec:combination}

\begin{figure*}[h]
    \centering
    \includegraphics[width=0.95\linewidth]{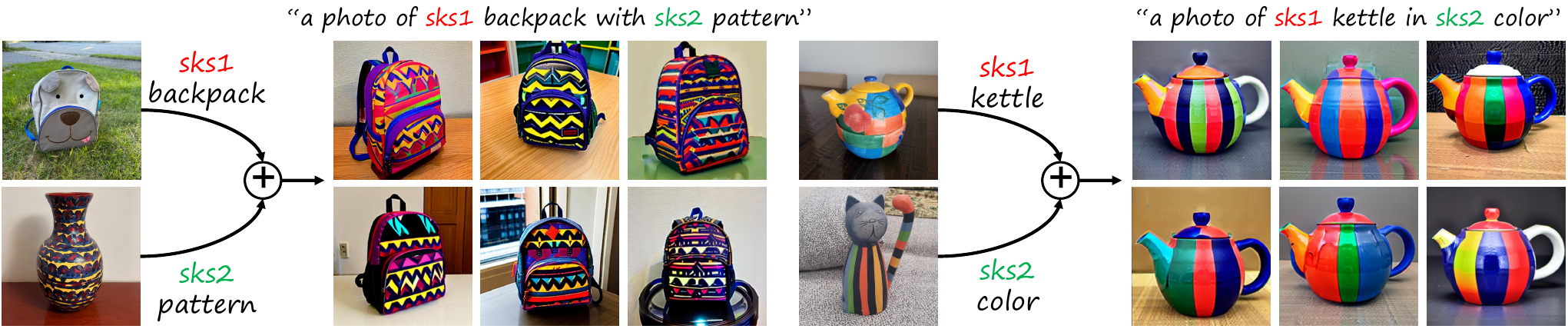}
    \caption{Given two different reference images, U-VAP can combine specific attributes from each concept and generate a new concept. Left: the combination of the pattern of the red vase and the specific backpack. Right: the combination of the color of the cat statue and the specific kettle.}
    \label{fig:combination}
\end{figure*}

We demonstrated the ability of U-VAP to combine different learned attributes.
As shown in Fig.~\ref{fig:combination}, given two different reference concepts, U-VAP can combine their different attributes.
For example, when given an image of a vase with a unique pattern and a backpack with a dog face pattern, we construct attribute-aware samples for each to learn the pattern of the former and the structure of the latter.
Subsequently, U-VAP simultaneously learns two target attributes on the same model and represents them with different identifiers (``sks1'' and ``sks2'').

In the inference phase, the user could directly write them in a prompt (as shown in Fig.~\ref{fig:combination}: ``a photo of sks1 pattern sks2 backpack'') and generate a new concept.
This new concept has the unique serrated pattern of the reference vase and the backpack's structure.
Similarly, given a colored cat statue and an image of a colored kettle, we generate a new concept of an appearance with colored stripes and a kettle structure that is the same as the reference concept.

\begin{figure*}[t]
    \centering
    \includegraphics[width=0.90\linewidth]{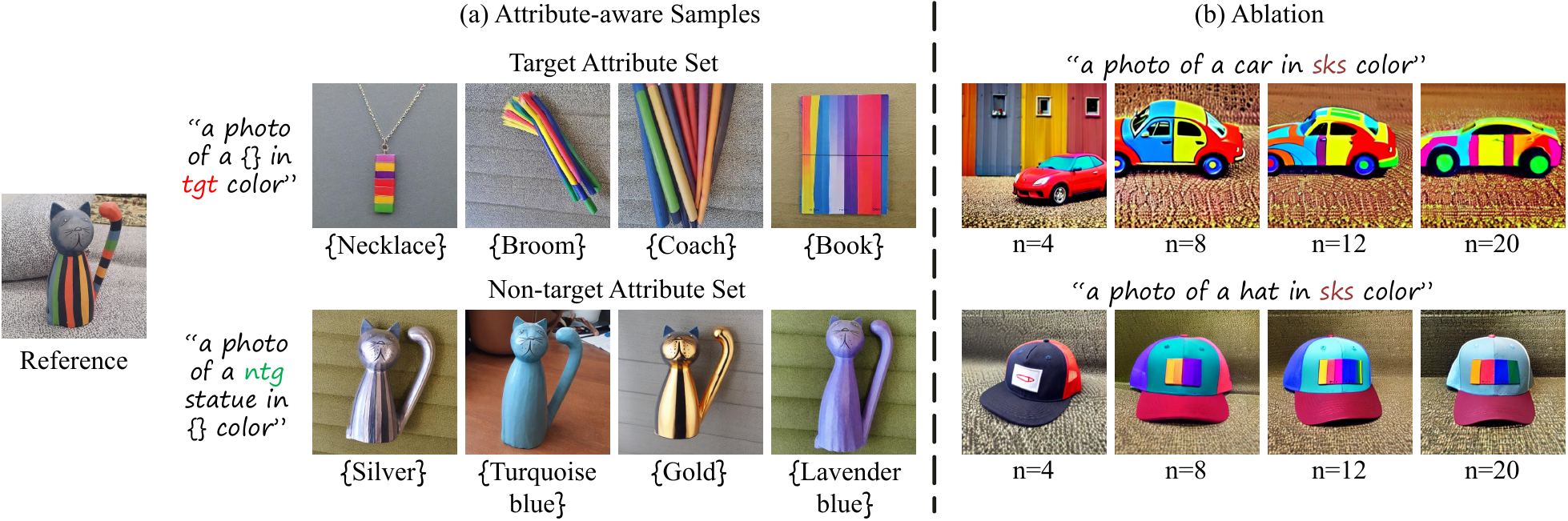}
    \caption{When specifying the color of the cat statue for personalization, the constructed attribute-aware samples are shown in (a). The target attribute set comprises various novel concepts with the reference color of the reference cat statue, while the non-target attribute set includes cat statues with different appearances. They are generated with target and non-target descriptions respectively. (b) demonstrates the generated results of various new concepts with the specified color using different quantities of attribute-aware samples. As the number of attribute-aware samples increases, the generated images more accurately embody the specified color.}
    \label{fig:ablation_sup}
\end{figure*}

\section{Attribute-aware Samples}
\label{sec:Data}

Fig.~\ref{fig:ablation_sup} (a) illustrates a pair of attribute-aware sets constructed from the reference image and the target attribute by U-VAP. 
Specifically, when the target attribute is ``color'', different objects in the target attribute set exhibit the same color as the reference concept. 
We obtain the target attribute set by conditioning the initial concept-aware model with a target description (``a photo of tgt color {}'')
Conversely, in the non-target attribute set generated by the non-target description, the concepts in each image have colors different from the reference concept.
Although images in the target attribute set have the specified color of reference concept, some of them are not in line with the description entirely.
For example, the initial model fails to generate ``broom'' or ``coach'' correctly, so we can not use it directly as a great attribute-aware model. 
Nonetheless, we only care about the target attribute (color of reference statute) of the target attribute set, and the error of other non-target attributes does not affect the learning of target identifier ``tgt''.

Using these constructed samples, we evaluate the influence of the number of attribute-aware samples on specific appearance learning, as shown in Fig.~\ref{fig:ablation_sup} (b). 
We generated a car, sunglasses, and a hat using the statue's color in the reference image, represented by ``sks''. 
Taking the example of ``hat'', when the number of attribute-aware samples (represented as n) is 4, the generated results barely match the reference concept's color. 
However, when n=8, the generated results meet the requirements. 
The generated concept has the same colorful stripe as the reference.
Further increasing n, we observed minimal improvement in learning the specified appearance.

\section{Limitation and Discussions}
\label{sec:limitation}
\begin{figure}[!t]
    \centering
    \includegraphics[width=0.95\linewidth]{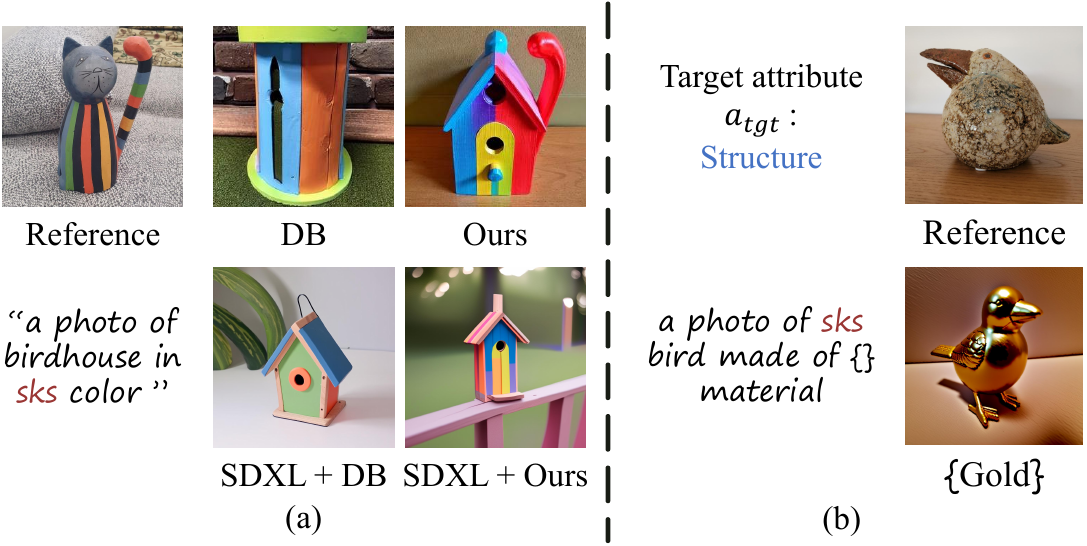}
    \caption{Bad cases. 
    (a) DreamBooth (DB)~\cite{ruiz2023dreambooth} and U-VAP fails in generation of some concept such as ``birdhouse'' because of the limitation of SD-based personalization. Based on a better basic model like SDXL, U-VAP overcomes this limitation and achieves better decoupling (SDXL + Ours).
    (b) After personalizing the structure of the bird statue and bounding it with the identifier ``sks'', U-VAP could generate bad results because of the influence of  the prior information
    of certain words (``bird'').}
    \label{fig:badcase}
\end{figure}
\begin{figure*}[!htbp]
    \centering
    \includegraphics[width=0.85\linewidth]{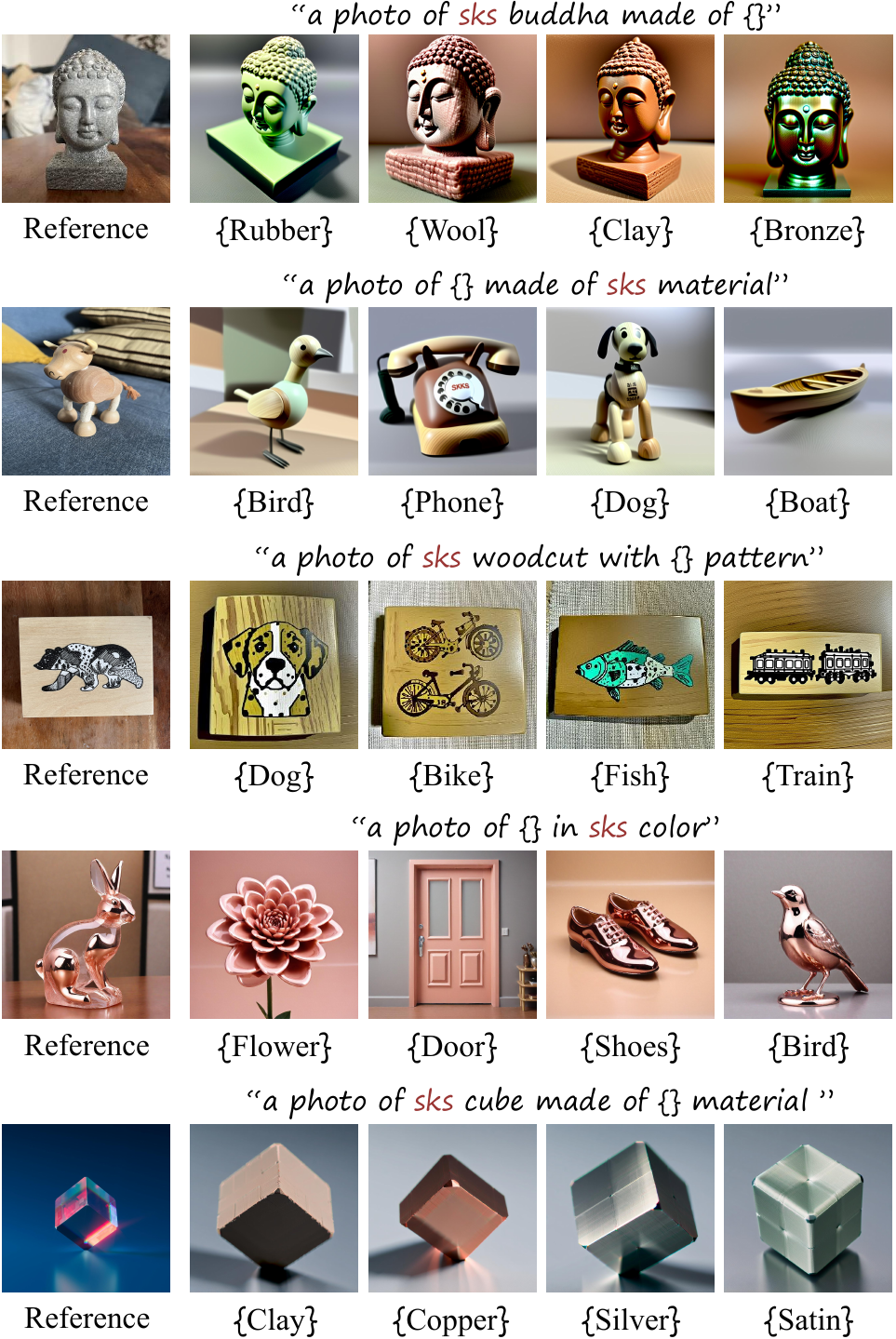}
    \caption{Results by SDXL-based U-VAP.}
    \label{fig:result_sdxl}
\end{figure*}
Because of the dependence on the capability of basic concept-aware personalization methods (such as DreamBooth (DB)~\cite{ruiz2023dreambooth}) in the pre-learning step, U-VAP could have poor performance in decoupling attributes of some cases, as shown in Fig.~\ref{fig:badcase} (a).
Given a photo of the colorful cat statue, we compare DB and U-VAP in generating a birdhouse in the specific color based on two different models, respectively.
The original DreamBooth and U-VAP are based on Stable Diffusion (SD) v1.5~\cite{rombach2022high}.
Additionally, we utilize SDXL~\cite{podell2023sdxl} in DreamBooth and our method for further comparison.  
Although the original U-VAP performs better than DreamBooth, it fails to completely decouple the undesired tail of the cat statue.
DreamBooth on SD v1.5 has limited ability to accurately decouple the target attribute, leading to low-quality constructed attribute-aware samples and bad results at the end.

However, based on the SDXL model, which performs better capability in text-to-image generation compared to previous versions of Stable Diffusion, the results of both DreamBooth and U-VAP achieve better quality.
We believe that with a better basic model like SDXL, the ability of attribute personalization of DreamBooth is improved and the efficiency of U-VAP can be further increased with high-quality attribute-aware samples.



In Fig.~\ref{fig:result_sdxl}, we provide more results generated by SDXL-based U-VAP to demonstrate the quality of specified appearance personalization.

Furthermore, in Fig.~\ref{fig:badcase} (b) we show another type of bad case where the structure of ``bird'' in the inference prompt suppresses the target attribute ``structure'' of the reference image in the result.
This is because the prior information of certain words may have stronger dominance in semantics and influence the learned attributes in some cases.

 
\end{document}